\documentclass[accepted]{uai2026} 
                        
\usepackage[american]{babel}
\usepackage{multirow}
\usepackage{natbib} 
\usepackage{mathtools} 
\usepackage{booktabs} 
\usepackage{tikz} 

\usepackage{times}
\usepackage{latexsym}

\usepackage[T1]{fontenc}

\usepackage[utf8]{inputenc}

\usepackage{microtype}

\usepackage{tcolorbox}
\tcbuselibrary{skins, breakable}
\usepackage{amsmath,amsfonts}
\usepackage{algorithmic}
\usepackage{algorithm}
\usepackage{array}
\usepackage[caption=false,font=normalsize,labelfont=sf,textfont=sf]{subfig}
\usepackage{textcomp}
\usepackage{stfloats}
\usepackage{amssymb, booktabs}
\usepackage{url}
\usepackage{verbatim}
\usepackage{graphicx}
\usepackage{cite}
\usepackage{multirow}
\usepackage[utf8]{inputenc}
\usepackage{amsmath}
\usepackage{amssymb}
\usepackage{geometry}

\usepackage{inconsolata}

\usepackage{graphicx}

\title{Not All Queries Need Deep Thought: CoFiCot for Adaptive Coarse-to-fine Stateful Refinement}

\author{
  \textbf{Dongxu Zhang\textsuperscript{1,$\ast$}}, \textbf{Hongqiang Lin\textsuperscript{2,$\ast$}}, \textbf{Yiding Sun\textsuperscript{1,$\ast$}}, \textbf{Pengyu Wang\textsuperscript{3},} \\
  \vspace{-10pt} 
  \textbf{Qirui Wang\textsuperscript{1},}  \textbf{Ning Yang\textsuperscript{4}}, \textbf{Jihua Zhu\textsuperscript{1},}  \\
  \vspace{10pt} 
  \textsuperscript{1}Xi’an Jiaotong University \quad
  \textsuperscript{2}Zhejiang University \\
  \textsuperscript{3}The Chinese University of Hong Kong (Shenzhen) \quad
  \textsuperscript{4}Institute of Automation, CASIA \\
  \vspace{10pt}
  \texttt{zhangdongxu@stu.xjtu.edu.cn}
}

\newcommand{\LINECOMMENT}[1]{\STATE \textit{// #1}}
\newcommand{\INLINECOMMENT}[1]{\quad \textit{// #1}}
  
\begin{document}
\maketitle

\begingroup
\renewcommand{\thefootnote}{}
\footnotetext{\textsuperscript{$\ast$}Equal contribution.}
\endgroup

\begin{abstract}

Scaling test-time computation enhances LLM reasoning ability but faces a uniform computation paradox. Allocating identical resources leads to over-correction on simple tasks and insufficient refinement on complex ones. To address this, we propose CoFiCot, a coarse-to-fine adaptive framework that dynamically tailors inference strategies to problem difficulty. Specifically, we implement a multi-metric classifier that triages queries by synthesizing semantic entropy, consensus reliability, and predicted reasoning depth . This enables a differentiated refinement stage that applies efficient aggregation for simple queries while routing complex ones to a context-aware correction loop . We formalize correction as a stateful sequential propagation process , where each repair is strictly conditioned on the verified history  of prior rectifications. By integrating Process Reward Models (PRMs)  within this state-dependent trajectory, CoFiCot effectively bridges the gap between granular error localization and global logical coherence, preventing the context fragmentation  typical of stateless refinement methods.

\end{abstract}

\section{Introduction}

The advent of Chain of Thought (CoT) prompting has endowed Large Language Models (LLMs) with the ability to deconstruct complex problems into intermediate steps~\citep{kojima2022large,zhang2026chain}.
Following this trajectory, recent advances like OpenAI's o1~\citep{jaech2024openai} and DeepSeekR1~\citep{guo2025deepseek} have empirically validated the test-time scaling law, demonstrating that allocating additional compute to extended reasoning trajectories yields performance gains comparable to parameter scaling. As a result, optimizing test-time computation has emerged as the pivotal frontier for next generation reasoning~\citep{snell2024scaling}. 
Methods such as Self-Consistency~\citep{wang2022self} and Best-of-$k$~\citep{sun2024easy} leverage this by generating and aggregating diverse reasoning paths.

However, allocating identical computational resources to every query is fundamentally misaligned with the variable complexity of reasoning tasks, as real world problems demand diverse levels of cognitive effort~\citep{wang2023math,yang2025decision}. 
\begin{figure}[t]
  \centering
  \includegraphics[width=0.95\linewidth]{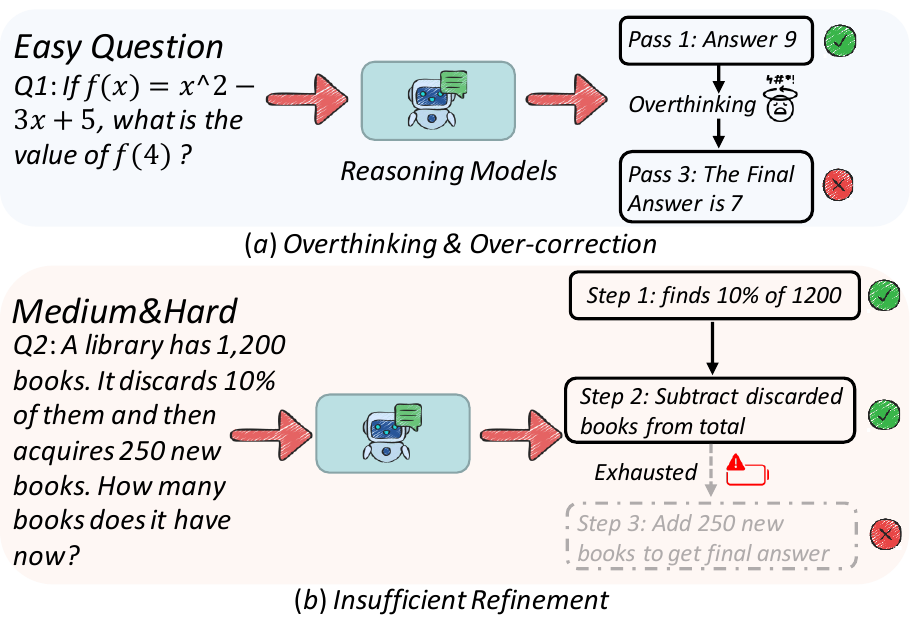}
  \caption{Conceptual illustration of the fundamental paradox of uniform computation in LLM reasoning. (Top) For an \textit{Easy} Question, the model correctly computes the answer in its initial pass but is forced into unnecessary iterations. This overthinking leads to over correction, where the correct answer is corrupted into a final incorrect one. (Bottom) For a \textit{Medium\&Hard} Question, the same fixed computational budget is insufficient. The reasoning process is prematurely terminated before all logical steps are completed, resulting in an insufficient refinement failure.}
  \label{figure1}
\end{figure}
This misalignment manifests differently across difficulty levels. For simple tasks, forcing extended reasoning not only causes rapid performance saturation~\citep{wadhwa2024learning}, but also leads to hallucination. For complex tasks, this uniform strategy fails to address the inherent brittleness of reasoning chains, where a single error can invalidate the entire process~\citep{lightman2023let}. In light of this, current LLMs' inability to emulate this adaptive allocation results in the critical uniform computation paradox, characterized by the dual failure modes of over thinking simple queries and under refining complex ones, as illustrated in Figure~\ref{figure1}.

To resolve the paradox of uniform allocation, the field is shifting towards adaptive computation~\citep{zhang2025ascot,huang2023large}, aiming to tailor inference workflows dynamically. However, current approaches are often specialized, failing to bridge the gap between efficiency and robust error recovery. Although routing strategies~\citep{xia2025tokenskip} improve efficiency, they lack the mechanisms to actively repair reasoning errors. Furthermore, iterative refinement methods~\citep{madaan2023self} often function in a stateless manner, leading to the phenomenon where correcting an intermediate step invalidates the subsequent logical flow.

Drawing inspiration from effective human problem solving, which relies on metacognitive triage to allocate minimal resources to straightforward tasks while reserving iterative effort for complex ones~\citep{zhang2026igasa,sun2026alignadaptrethinkingparameterefficient}, we introduce CoFiCot, a Coarse-to-fine framework designed to resolve the paradox of uniform computation. Our method operates via a two stage pipeline. First, a coarse-grained classification stage acts as a lightweight semantic router which triages problems into \textit{Easy}, \textit{Medium}, or \textit{Hard}. Second, a fine-grained refinement stage executes a differentiated strategy. \textit{Easy} problems are resolved via efficient aggregation, while \textit{Medium} and \textit{Hard} problems enter an iterative correction loop for further refinement.

Crucially, we introduce a Stateful Sequential Correction mechanism designed to enforce logical consistency during the repair process. Unlike prevailing methods that regenerate entire chains~\citep{madaan2023self}, our mechanism formalizes correction as a history generative process. By anchoring the validated reasoning path and initiating a new decoding branch from the point of error, we ensure that corrections trigger a cascading update of all dependent reasoning steps, reconciling the precision of step-level error localization with the coherence required for complex reasoning.

Our contributions are summarized as follows:
\begin{itemize}
\item We propose CoFiCot, an adaptive framework that dynamically matches reasoning strategies according to the problem difficulty. 
\item We devise a sequential correction mechanism for complex problems. Our approach treats reasoning as a state-dependent trajectory, ensuring corrections propagate downstream.
\item Extensive experiments across seven benchmarks demonstrate that CoFiCot outperforms strong baselines and achieves a better trade-off between accuracy and efficiency.
\end{itemize}

\section{Related Work}
\subsection{Enhancing Reasoning Ability in LLMs}

Recent research in LLM reasoning has shifted focus from optimizing CoT prompting~\citep{kojima2022large} to ensemble based inference strategies~\citep{zhang2026pointcot}. To mitigate the fragility of single reasoning paths, aggregation strategies like Self-Consistency~\citep{wang2022self} and Best-of-$k$~\citep{lightman2023let} generate multiple candidates to select a consensus or high reward answer. However, these methods suffer from performance saturation, where computational costs rise linearly while accuracy gains plateau~\citep{li2024confidence,yin2024aggregation}.

Parallel to these efforts, Iterative Refinement has emerged as a promising paradigm, allowing models to actively improve outputs~\citep{madaan2023self}. Yet, indiscriminate refinement causes over correction on simple problems, corrupting correct answers~\citep{chen2024not,liu2024large} while providing insufficient rectification for complex reasoning failures~\citep{li2024confidence}. These challenges underscore that simply iterating is insufficient.

\subsection{Computational Efficiency in LLMs}

The high computational cost of CoT has driven research into efficiency via compression techniques like token pruning~\citep{xia2025tokenskip} or step reduction~\citep{kumar2025overthink}. While these methods mitigate per-token costs, they often remain limited by static inference paths that fail to account for varying query complexities \citep{schwartz2020right}. A more promising paradigm is Adaptive Computation, where Coarse-to-fine (C2F) frameworks attempt to tailor resource allocation dynamically~\citep{wang2023enhancing}. However, current C2F solutions often face a trade-off between latency and corrective capability. Methods like MAgICoRe~\citep{chen2024magicore} incur prohibitive latency, whereas early exit strategies~\citep{zhang2025ascot} lack mechanisms to actively repair flawed reasoning. CoFiCot bridges this gap by unifying granular difficulty assessment with stateful refinement to achieve an optimal balance between accuracy and efficiency.

\subsection{Feedback-Driven Error Rectification}

The efficacy of iterative refinement hinges on feedback quality \citep{chen2024not, wang2022self}. Intrinsic self-correction often fails, as models struggle to self-diagnose errors and are prone to generating invalid solutions \citep{cao2025large, yang2025token}. Accordingly, research has pivoted toward external feedback via specialized Reward Models (RMs) \citep{li2024confidence}. Although Outcome RMs (ORMs) evaluate entire reasoning traces \citep{ma2025step, lu2024llm}, Process RMs (PRMs) provide dense, step-level signals \citep{zhu2025breaking}, current integration methods often results in a lack of global coherence. CoFiCot addresses this limitation by formalizing correction as a propagation process, ensuring that RM-guided repairs maintain the integrity of the entire reasoning.

\begin{figure*}[t]
  \centering
  \includegraphics[width=1.0\linewidth]{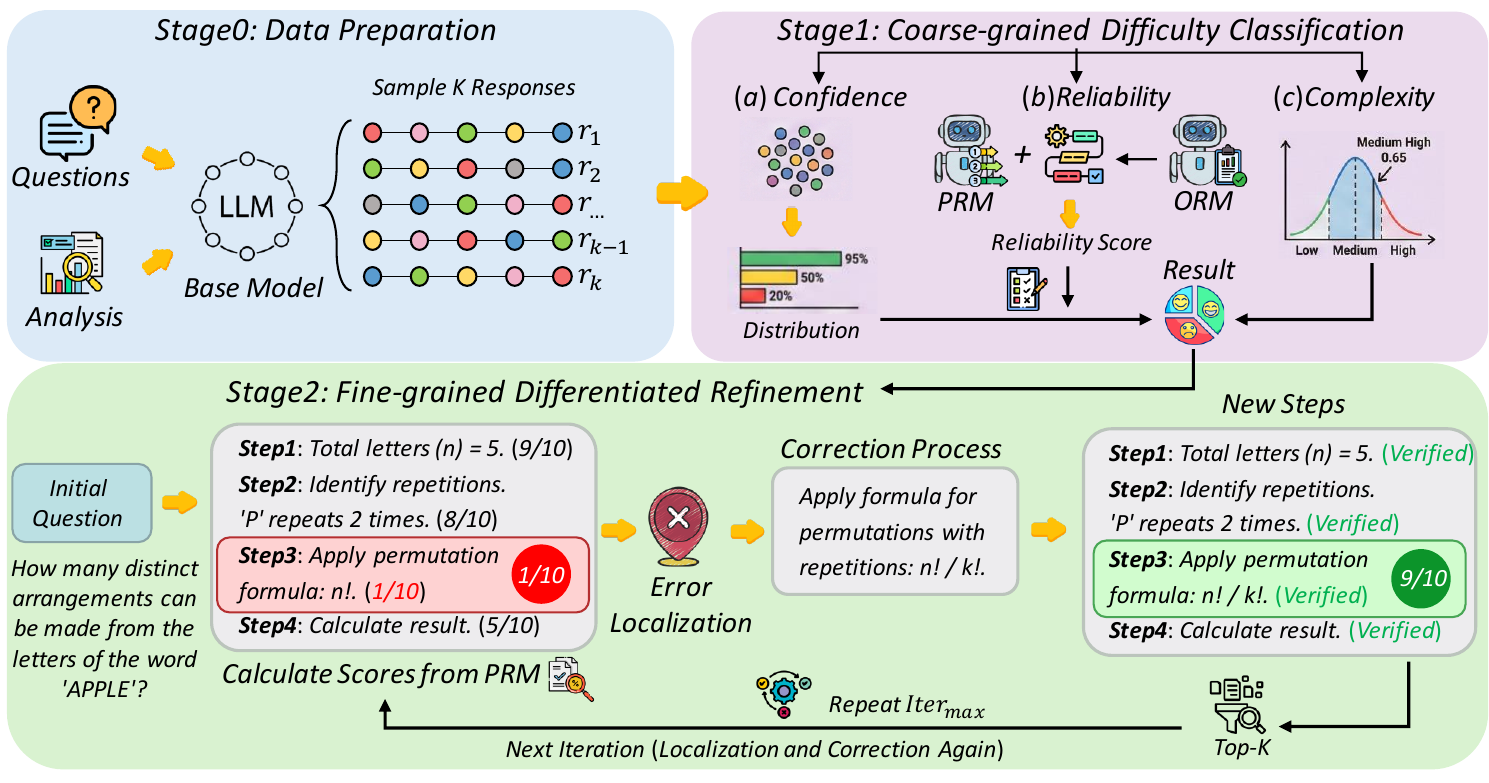}
  \caption{The complete workflow of the CoFiCot framework. The process begins in Stage 0, where the base LLM generates an initial ensemble of $k$ reasoning traces. This set is passed to Stage 1, which performs a parallel, multi metric analysis to assess difficulty. \textit{Easy} problems are resolved with simple aggregation. \textit{Medium} and \textit{Hard} problems are channeled into Stage 2. This stage initiates an iterative loop. A PRM scores each step, identifying a flawed step, which is then fed into the Correction Process. Critically, this correction is context aware, conditioning on the history of previous, correct steps to generate a new solution. The refined solution is then selected via an ORM, and the process repeats until termination criteria are met.}
  \label{figure2}
\end{figure*}

\section{Methodology}
\subsection{Overview and Data Preparation}

CoFiCot operates through a tiered inference pipeline that begins by constructing a diverse solution space to facilitate subsequent complexity analysis. During the data preparation stage, we initialize the workflow by generating an ensemble of $k$ candidate solutions sampling from a base LLM. The resulting set denoted as $\mathcal{R}_0 = \{r_1, \dots, r_k\}$, consists of independent CoT reasoning traces. Each solution $r_i \in \mathcal{R}_0$ represents a self contained attempt at solving the problem. This solution set serves as the foundation for the Coarse-to-fine procedure, where the coarse-grained classification stage acts as an analytical routing mechanism to evaluate problem difficulty. Based on this assessment, queries are either resolved via immediate aggregation or channeled into a fine-grained refinement stage featuring an iterative correction loop for more challenging cases. The complete workflow is illustrated in Figure~\ref{figure2}, with formal execution details provided in Algorithm~\ref{alg:coficot} in the Appendix~\ref{sec:appendix}.

\subsection{Stage 1: Coarse-grained Classification}
This stage performs a systematic analysis of the initial solution set $\mathcal{R}_0$ to classify each problem into one of three categories: \textit{Easy}, \textit{Medium} or \textit{Hard}. The classification is derived from a synthesis of three complementary metrics.

\textbf{Confidence Assessment}\quad This metric gauges predictive uncertainty via the semantic consensus of $\mathcal{R}_0$. We partition the generated solutions into $n$ semantic clusters $\mathcal{A} = \{\mathcal{A}_1, \dots, \mathcal{A}_n\}$ and define a reward probability for the $i$-th cluster:
\begin{equation}
    p(\mathcal{A}_i) = \frac{\sum_{r_j \in \mathcal{A}_i} \mathcal{S}_j}{\sum_{m=1}^{n} \sum_{r_l \in \mathcal{A}_m} \mathcal{S}_l},
\end{equation}
where $\mathcal{S}_j$ denotes the RM score of solution $r_j$. We quantify the predictive uncertainty via the Shannon entropy $H(\mathcal{A})$ of the answer cluster distribution:
\begin{equation}
    H(\mathcal{A}) = -\sum_{i=1}^{n} p(\mathcal{A}_i) \log p(\mathcal{A}_i).
\end{equation}

To facilitate thresholding, we invert and normalize $H(\mathcal{A})$ into a confidence score $C \in [0, 1]$ via a scaled sigmoid function:
\begin{equation}
    C = \sigma(\alpha \cdot (1 - H(\mathcal{A}))),
\end{equation}
where $\alpha$ is a scaling hyperparameter. The preliminary difficulty label $d_1$ is then assigned based on empirical thresholds:
\begin{equation}
    d_1 = \begin{cases}
    \text{\textit{Easy}} & \text{if } C > 0.6 \\
    \text{\textit{Medium}} & \text{if } 0.3 < C \le 0.6 \\
    \text{\textit{Hard}} & \text{if } C \le 0.3
    \end{cases}.
\end{equation}

The low entropy typically signifies a distribution sharply peaked around a consensus, we acknowledge that high confidence does not guarantee correctness due to potential confident hallucination. Therefore, this metric serves as a preliminary filter and is coupled with the Reliability to validate the consensus quality (detailed in Appendix~\ref{app:stage1_details}).

\textbf{Reliability Assessment}\quad This metric contextualizes the consensus quality to filter out false consensus scenarios. We identify the majority answer cluster $\mathcal{A}_g$, compute its average RM score $\bar{\mathcal{S}}_g$, and normalize it against the global score distribution of $\mathcal{R}_{0}$ via a Z-score transformation:
\begin{equation}
    \mathcal{S}_{\text{norm}} = \frac{\bar{\mathcal{S}}_g - \mu_{\text{all}}}{\sigma_{\text{all}}},
\end{equation}
where $\mu_{\text{all}}$ and $\sigma_{\text{all}}$ represent the global mean and standard deviation. A negative $\mathcal{S}_{\text{norm}}$ indicates that the popular consensus underperforms the candidate pool's average quality. Accordingly, we map this score to the second difficulty label $d_2$:
\begin{equation}
    d_2 = \begin{cases}
    \text{\textit{Easy}} & \text{if } \mathcal{S}_{\text{norm}} \ge \delta \\
    \text{\textit{Medium}} & \text{if } 0 \le \mathcal{S}_{\text{norm}} < \delta \\
    \text{\textit{Hard}} & \text{if } \mathcal{S}_{\text{norm}} < 0 \\
    \end{cases},
\end{equation}
where $\delta$ is a positive threshold that delimits high quality consensus. This ensures that answers with significant quality superiority are trusted as \textit{Easy}.

\textbf{Complexity Assessment}\quad Finally, we estimate the intrinsic problem complexity by prompting the base LLM to predict the necessary reasoning steps:
\begin{equation}
N_{\text{steps}} = \mathcal{G}_{\text{llm}}(Q \oplus P_{\text{predict}}).
\end{equation}

This step incurs negligible overhead as it requires generating only a single integer token. To determine the difficulty label $d_3$, we map $N_{\text{steps}}$ against the inverse CDF $F_L^{-1}$ of solution lengths from a representative corpus, utilizing tertile based thresholds (details provided in Appendix~\ref{app:stage1_details}):
\begin{equation}
d_3 =\begin{cases}\text{\textit{Easy}} & \text{if } N_{\text{steps}} \le F_L^{-1}(0.33) \\
\text{\textit{Medium}} & \text{else }  \\
\text{\textit{Hard}} & \text{if } N_{\text{steps}} > F_L^{-1}(0.67)\end{cases}.
\end{equation}

\textbf{Final Difficulty Synthesis}\quad To obtain the final classification, we employ a Balanced Strategy. The final difficulty score is a weighted sum of the mapped metric values:
\begin{equation}
    D_{\text{score}} = \sum_{i=1}^{3} w_i \cdot \text{val}(d_i),
\end{equation}
where $d_i$ denotes the difficulty label assigned by the $i$-th metric, $\text{val}(\cdot)$ represents the mapping function that transforms each qualitative difficulty label into a discrete numerical rank and $w_i$ is the scalar weight reflecting the relative importance of each metric. We provide a detailed justification for these hyperparameters and an analysis of their sensitivity in Appendix~\ref{app:stage1_details}.

\subsection{Stage 2: Fine-grained Refinement}
Upon determining the difficulty of the problem $D_{\text{final}}$, the framework executes a differentiated refinement strategy. 
\textit{Easy} problems bypass the expensive refinement loop and are resolved on the initial ensemble $\mathcal{R}_0$ (details in Appendix~\ref{app:refinement}).

Problems classified as \textit{Medium} or \textit{Hard} are subjected to an iterative refinement loop. Let $\mathcal{R}_{t-1}$ be the set of $k$ candidate solutions at the beginning of iteration $t$. The refinement process employs a sequential correction mechanism, modeled as a state-dependent transformation function $\Phi$. 
For each solution $r_i \in \mathcal{R}_{t-1}$ composed of steps $\{s_{i,1}, s_{i,2}, \dots\}$, a PRM provides step-level scores. Steps falling below a threshold $\tau_{\text{step}}$ are flagged as errors. Upon detecting the first error step $s_{i,j}$, we freeze the verified history $H_{i, j-1}^{(t)}$ and invoke $\Phi$ (details in Appendix~\ref{app:stage1_details}) to generate a corrected step. 
To ensure the correction logically propagates, the generation is conditioned on the original question $Q$ and the history:
\begin{equation}
    s_{i,j}^{(t)} = \Phi\left(Q, s_{i,j}^{(t-1)}, \mathcal{F}_{i,j}, H_{i, j-1}^{(t)}\right),
\end{equation}
where $Q$ is the problem statement, $s_{i,j}^{(t-1)}$ is the original erroneous step, $\mathcal{F}_{i,j}$ is the PRM derived feedback, and $H_{i, j-1}^{(t)} = \{s_{i,1}^{(t)}, \dots, s_{i,j-1}^{(t)}\}$ is the sequence of verified steps preceding the error. 

Crucially, unlike stateless methods that simply replace the error step, our mechanism adheres to causal consistency. Verified steps before the error are carried over unchanged ($s_{i,k}^{(t)} = s_{i,k}^{(t-1)}$ for $k < j$), while all steps following the correction are regenerated based on the updated state. The refined solution $r_i^{(t)}$ is thus a coherent sequence assembled from the preserved history and the corrected step.

Following the correction of all $k$ solutions, an iterative selection and evaluation process is employed to curate the most promising candidates for the subsequent iteration. Let $\mathcal{R}^{(t)} = \{r_1^{(t)}, \dots, r_k^{(t)}\}$ be the set of $k$ newly refined solutions. This set is merged with the original set $\mathcal{R}_{t-1}$ to form an expanded candidate pool:
\begin{equation}
\mathcal{P}_t = \mathcal{R}_{t-1} \cup \mathcal{R}^{(t)}.
\end{equation}

An ORM then provide a quality score for each solution in this pool. The top-$k$ solutions that maximize the aggregate quality are selected to form the input set for the next iteration $\mathcal{R}_t$, by solving the following optimization problem:
\begin{equation}
    \mathcal{R}_t = \underset{\mathcal{S} \subseteq \mathcal{P}_t, |\mathcal{S}|=k}{\arg\max} \sum_{r \in \mathcal{S}} \text{ORM}(r).
\end{equation}

This iterative cycle of correction and selection is governed by a dynamic termination protocol executed at the end of each iteration. The expanded solution set $\mathcal{R}_t$ is reevaluated by the Stage 1 classifier. We also employ an Early Exit strategy. If the difficulty downgrades to \textit{Easy}, the loop terminates immediately to save compute. Otherwise, the refinement continues until the predefined budget ($Iter_{max}$) is exhausted (details in Appendix~\ref{app:refinement}).

\section{Experiments}
\subsection{Experimental Setup}
We outline the experimental setup spanning seven diverse benchmarks and two backbone models. Furthermore, we compare against strong baselines to rigorously evaluate the trade-off between accuracy and efficiency of our CoFiCot.

\textbf{Datasets and Evaluation Metrics}\quad
We conduct our evaluation across seven challenging datasets, which can be grouped into mathematical reasoning and general reasoning~\citep{wang2025multimodal}.
We use a diverse suite of five reasoning benchmarks. This includes GSM8K~\citep{cobbe2021training} and SVAMP~\citep{patel2021nlp}, which are math word problems; MATH~\citep{hendrycks2021measuring}, a benchmark of high school competition level mathematics problems; and MMLU~\citep{hendrycks2020measuring} and SAT~\citep{zhong2023agieval}, which are standard mathematical reasoning subsets.
To test the generalization of our framework beyond mathematical domains, we also evaluate on two additional tasks: ARC~\citep{clark2018think}, a complex commonsense reasoning dataset and Date~\citep{srivastava2023beyond}, which requires logical reasoning about dates.
Our evaluation is based on two primary categories of metrics. The primary metric is the final accuracy of the model. And we also report the average number of samples generated per question denoted as $k$, which accounts for the adaptive nature of our refinement loop and the total number of decoded tokens.

\begin{table*}[t]
  \centering
   \resizebox{0.85\textwidth}{!}{ 
  \begin{tabular}{lcccccc}
    \toprule
  \textbf{Method} & \textbf{MMLU} & \textbf{MATH} & \textbf{SVAMP} & \textbf{GSM8K} & \textbf{SAT} & \textbf{Avg.} \\
    \midrule
    \multicolumn{7}{c}{\textit{Llama3-8B-Instruct}} \\
    \midrule
    Zero-shot CoT~\citep{wei2022chain} & 50.4 & 24.2 & 72.4 & 80.1 & 58.2 & 57.1 \\
    Self-Refine~\citep{madaan2023self} & 49.6 & 24.6 & 72.0 & 79.0 & 57.7 & 56.6 \\
     Best-of-$k$ ($k$ = 120)~\citep{lightman2023let} & 62.6 & 41.4 & 88.7 & 90.1 & 72.4 & 71.0 \\
     $k$-way SC ($k$ = 120)~\citep{wang2022self} & 63.0 & 40.6 & 89.8 & 90.3 & 70.5 & 70.8 \\
     Self-Refine + $k$-way SC~\citep{chen2024magicore} & 62.1 & 40.4 & 88.6 & 90.1 & 68.2 & 69.9 \\
     CoFiCot(ours) & \textbf{68.8} & \textbf{47.9} & \textbf{91.4} & \textbf{91.8} & \textbf{75.0} & \textbf{75.0} \\
    \midrule
    \multicolumn{7}{c}{\textit{GPT-3.5-Turbo}} \\
    \midrule
    Zero-shot CoT~\citep{wei2022chain} & 62.5 & 37.2 & 78.1 & 78.5 & 76.8 & 66.6 \\
    Self-Refine~\citep{madaan2023self} & 62.4 & 37.4 & 77.7 & 77.4 & 77.3 & 66.4 \\
     Best-of-$k$ ($k$ = 120)~\citep{lightman2023let} & 70.1 & 50.6 & 87.7 & 90.5 & 87.8 & 77.3 \\
     $k$-way SC ($k$ = 120)~\citep{wang2022self} & 70.4 & 51.2 & 86.9 & 89.8 & 87.6 & 77.2 \\
     Self-Refine + $k$-way SC~\citep{chen2024magicore} & 69.4 & 49.8 & 86.9 & 88.1 & 85.6 & 76.0 \\
     CoFiCot(ours) & \textbf{73.5} & \textbf{57.7} & \textbf{89.8} & \textbf{91.2} & \textbf{90.1} & \textbf{80.5} \\
    \bottomrule
  \end{tabular}
   } 
  \caption{Performance comparison of our proposed CoFiCot method against baseline methods on the Llama3-8B-Instruct and GPT-3.5-Turbo models. Results are reported across five reasoning benchmarks. Best performance is highlighted in bold. We report accuracy(\%).}
  \label{tab1}
\end{table*}

\textbf{Models and Implementation Details}\quad
All experiments are conducted using Llama-3-8B-Instruct~\citep{dubey2024llama} and GPT-3.5-Turbo~\citep{schulman2022chatgpt} to ensure our findings are generalizable.
Our framework leverages two pretrained RMs. 
We employ InternLM-7B as ORM and Math-Shepherd-7B as PRM. 
For all experiments, we generate $40$ initial candidate solutions using a decoding temperature of 0.8 to ensure diversity. For our coarse stage, we apply uniform weights ($w_1=w_2=w_3=1/3$) to the scores from our three classification metrics. For the fine stage, we set the step-level verification threshold $\tau_{step} = 0.5$, and the maximum number of iterations ($Iter_{max}$) is set to 2 for \textit{Medium} problems and 3 for \textit{Hard} problems.
For hyperparameters, we set the entropy scaling factor $\alpha=2$ and the Z-score threshold $\delta=0.5$.

\textbf{Baselines for Comparison}\quad
We compare CoFiCot against with other strong baselines. Vanilla Prompting~\citep{kojima2022large} is a standard zero-shot CoT prompt, which generates a single reasoning chain. Aggregation based Methods~\citep{yin2024aggregation} generate a large number of samples and aggregate them. We compare against $k$-way Self-Consistency~\citep{li2023making}, which uses a majority vote and Best-of-$k$~\citep{lightman2023let}, which uses the ORM to select the best solution. To establish a performance ceiling, these baselines are given a high budget of $k=120$ samples.

\subsection{Main Results}
We present a comprehensive empirical evaluation of CoFiCot against strong baselines across two primary domains: mathematical reasoning and general commonsense reasoning.

\begin{figure}[t]
  \centering
  \includegraphics[width=0.85\linewidth]{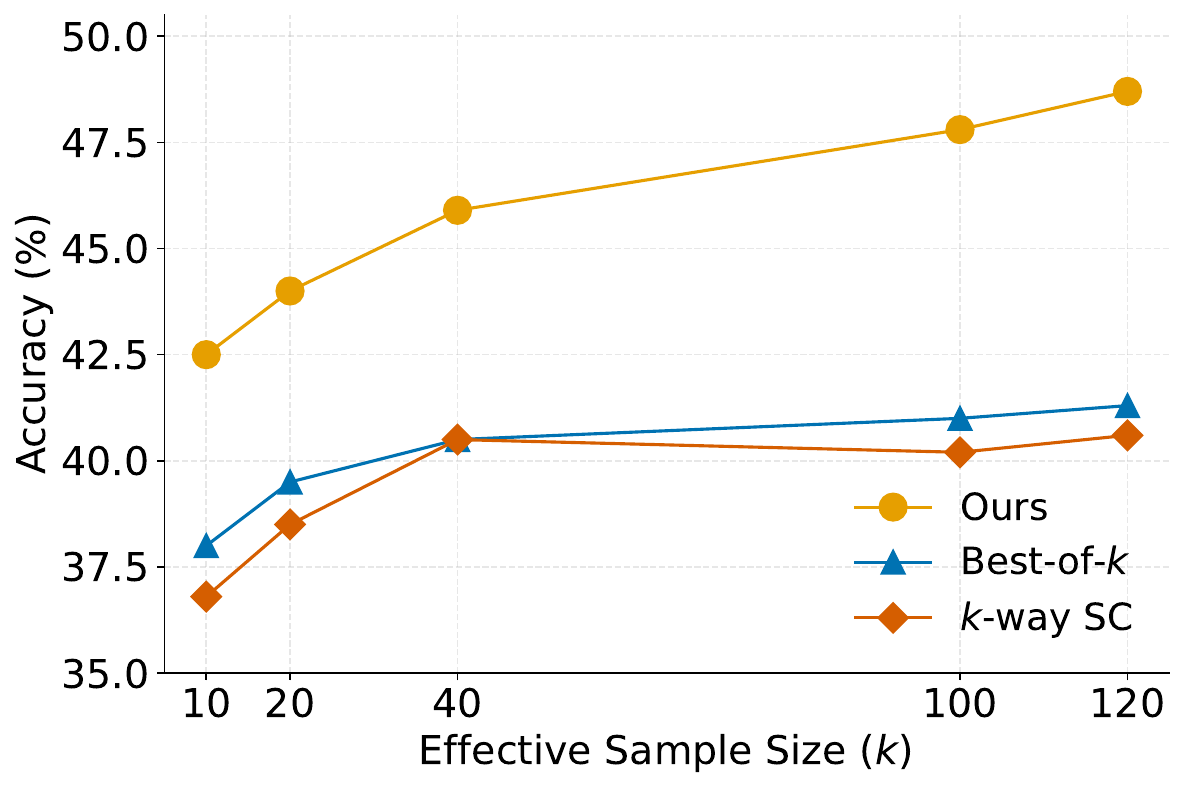}
  \caption{Accuracy vs. Effective Sample Size ($k$) on the MATH dataset.}
  \label{figure3}
\end{figure}

\textbf{Performance on Mathematical Reasoning}\quad
The primary results of our evaluation are presented in Table ~\ref{tab1}. The findings clearly demonstrate that CoFiCot consistently and significantly outperforms all baseline methods. On the Llama-3-8B-Instruct model, CoFiCot achieves an average accuracy of 75.0\%, representing a substantial 4.0\% absolute improvement over the strongest baseline Best-of-$k$. Notably, CoFiCot achieves a 6.5\% absolute gain over the baseline's 41.4\% on MATH. This trend is consistent on the GPT-3.5-Turbo model, where CoFiCot achieves an average accuracy of 80.5\%, surpassing the strongest baseline by 3.2\%. 

A hypothesis of our work is that brute force aggregation methods suffer from performance saturation. Figure~\ref{figure3} provides a clear visualization of this phenomenon on the MATH dataset. The accuracy of both $k$-way SC and Best-of-$k$ grows minimally as the sample size increases from $k=40$ to $k=120$, quickly hitting a performance plateau. In contrast, CoFiCot not only starts from a significantly higher baseline accuracy but also scales more effectively with an increased computational budget. Most notably, CoFiCot initiated with an effective sample size of $k=40$ achieves 45.9\% accuracy that is already superior to both $k$-way SC (40.6\%) and Best-of-$k$ (41.4\%) using a massive $k=120$ budget.

\begin{table}[t]
  \centering 
  \resizebox{0.80\columnwidth}{!}{
  \begin{tabular}{lcc}
    \toprule
    \textbf{Method} & \textbf{ARC} & \textbf{Date} \\
    \midrule
    Zero-shot~\citep{wei2022chain} & 66.5 & 52.5 \\
    40-way SC~\citep{wang2022self} & 85.5 & 72.5 \\
    120-way SC~\citep{chen2024magicore} & 86.0 & 72.5 \\
    CoFiCot(ours) & \textbf{88.2} & \textbf{80.8} \\
    \bottomrule
  \end{tabular}}
  \caption{Performance comparison on commonsense reasoning and logical reasoning benchmarks. We also report accuracy(\%).}
  \label{tab2}
\end{table}

\textbf{Generalization to Commonsense Reasoning}\quad
To demonstrate the general domain applicability of our framework, we evaluated CoFiCot on the ARC~\citep{clark2018think} and Date~\citep{srivastava2023beyond} datasets. As shown in Table ~\ref{tab2}, CoFiCot's superiority is not limited to mathematical tasks. Our method achieves an accuracy of 88.2\% in ARC, surpassing the baseline of SC 120-ways by 2.2\%. The improvement is more pronounced on the Date Understanding task, where CoFiCot (80.8\%) outperforms 120-way SC (72.5\%) by a significant 8.3\%.

\begin{table*}[t]
  \centering
  \renewcommand{\arraystretch}{1.1}
  \resizebox{0.85\textwidth}{!}{
  \begin{tabular}{lcccccc}
    \toprule
    \textbf{Method} & \textbf{MMLU} & \textbf{MATH} & \textbf{SVAMP} & \textbf{GSM8K} & \textbf{SAT} & \textbf{Avg.} \\
    \midrule
    Qwen2.5-Math-7B~\citep{yang2024qwen2} & 73.9 & 78.8 & 91.8 & 94.9 & 92.3 & 86.3 \\
    $k$-way SC ($k$ = 40)~\citep{wang2022self} & 81.3 & 87.0 & 95.5 & 97.2 & 97.3 & 91.7 \\
    $k$-way SC ($k$ = 120)~\citep{chen2024magicore} & 82.0 & 86.8 & 95.4 & \textbf{97.3} & 97.3 & 91.8 \\
    Best-of-$k$ ($k$ = 120)~\citep{lightman2023let} & 82.6 & 86.0 & 93.4 & 96.9 & 95.2 & 90.8 \\
    \midrule 
    CoFiCot(ours) & \textbf{84.9} & \textbf{91.7} & \textbf{95.8} & 97.2 & \textbf{97.6} & \textbf{93.4} \\
    \bottomrule
  \end{tabular}
  } 
\caption{Evaluating CoFiCot's scalability on the powerful Qwen2.5-Math-7B model. We report accuracy(\%).}
\label{tab3}
\end{table*}

\begin{table}[t]
  \centering 
  \begin{tabular}{ccc}
    \toprule
    \textbf{PRM} & \textbf{ORM} & \textbf{Acc.} \\
    \midrule
    Math-Shepherd-7B & InternLM-7B & 48.3 \\
    \midrule
    Qwen-Math 7B & InternLM-7B & 54.9 \\
    \midrule
    Math-Shepherd-7B & Llama-3.1-8B & 51.2 \\
    \bottomrule
  \end{tabular}
  \caption{Analysis of CoFiCot's sensitivity to the quality of its component reward models. Accuracy (Acc. \%) is shown for different PRM and ORM combinations.}
  \label{tab4}
\end{table}

\subsection{Additional Analysis}

To further probe the robustness and characteristics of our CoFiCot framework, we conduct a series of additional analyses covering scalability, modularity, and other critical aspects (detailed in Appendix~\ref{case study}).

\textbf{Performance Analysis with Different Models}\quad
A key question is whether CoFiCot's benefits are limited to improving weaker base models. To test this, we apply our framework to Qwen2.5-Math-7B, a model for mathematical reasoning. The results presented in Table ~\ref{tab3} demonstrate that CoFiCot continues to provide performance gains.

CoFiCot achieves an average accuracy of 93.4\%. The performance gains are particularly pronounced on the dataset MATH, where CoFiCot (91.7\%) substantially outperforms the baselines. This result confirms that CoFiCot is not merely a compensatory mechanism but an effective framework for scaling and enhancing the reasoning capabilities.

The performance of CoFiCot is guided by its external ORM and PRM. In Table~\ref{tab4}, we analyze the modularity of our framework by swapping these components on the MATH dataset. The results clearly indicate that CoFiCot's performance scales directly with the quality of its components.

Upgrading the PRM from Math-Shepherd-7B to the stronger Qwen-Math 7B yields the most substantial accuracy boost, from 48.3\% to 54.9\% (+6.6\%). Similarly, upgrading the ORM from InternLM-7B to Llama-3.1-8B also provides a significant gain (48.3\% to 51.2\%). This experiment validates the modular design of CoFiCot, demonstrating its ability to readily incorporate and benefit from advancements in reward models.

\begin{figure*}[t]
  \centering
  \includegraphics[width=0.95\linewidth]{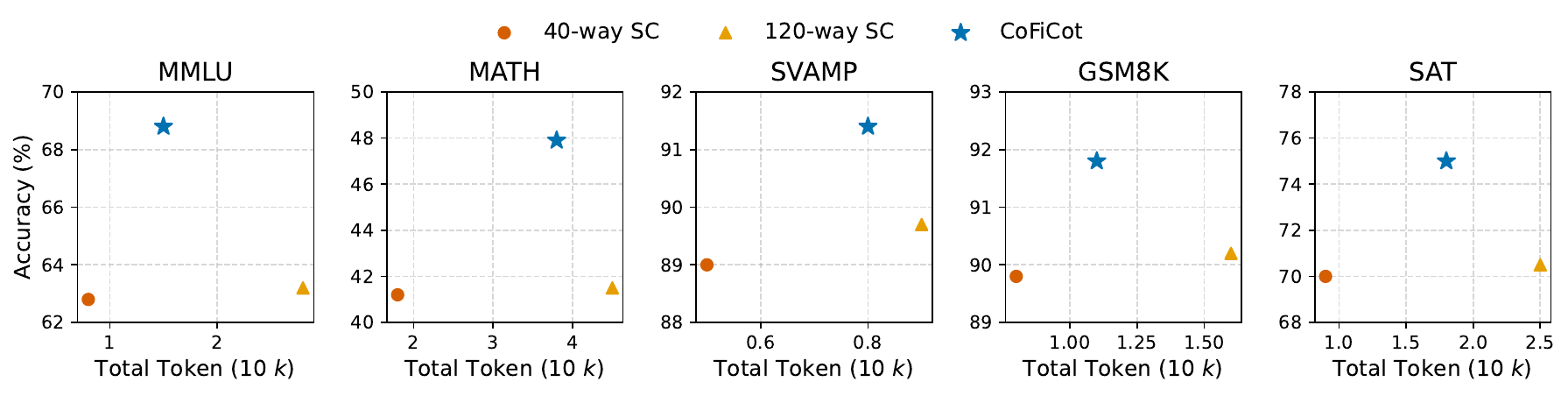}
  \caption{Token count and accuracy comparison across different datasets. The reported token count for CoFiCot includes the overhead from the initial sampling, Stage 1 classification tokens, and all PRM/ORM verification steps in Stage 2. While scaling Self-Consistency from $k=40$ to $k=120$ introduces substantial token overhead, our CoFiCot achieves significantly higher accuracy while being more token efficient than the 120-way SC baseline.}
  \label{figure4}
\end{figure*}

\textbf{Token Budget Analysis}\quad
While our main results demonstrated CoFiCot's superior sample efficiency, we provide a more granular analysis of the performance to cost trade-off in Figure~\ref{figure4}. This figure plots accuracy against the total normalized token consumption for CoFiCot and the SC baselines across all five reasoning datasets.

CoFiCot achieves higher accuracy while consuming fewer or comparable tokens than the computationally expensive 120-way SC baseline. On MATH, CoFiCot achieves 47.9\% accuracy using fewer tokens than 120-way SC. On MMLU and SAT, CoFiCot uses a fraction of the tokens consumed by 120-way SC to achieve a significantly higher accuracy. This analysis confirms that our adaptive method avoids the brute force cost of uniform sampling and provides a superior efficiency to performance ratio across the board.

\textbf{Latency vs. Trade-off}\quad
It is worth noting that while CoFiCot significantly reduces total token volume, the sequential nature of the Stage 2 refinement loop may incur higher latency per token compared to parallel sampling methods. Since CoFiCot bypasses the refinement loop for a large portion of \textit{Easy} queries (approx. 40-60\%), the average latency remains competitive. Furthermore, the reduction in computational load results in lower deployment costs and consumption (detailed in Appendix~\ref{case study}).

\begin{table}[t]
\centering
\small
\renewcommand{\arraystretch}{0.9}
\resizebox{0.8\linewidth}{!}{
\begin{tabular}{lcc}
\toprule
\multirow{2}{*}{\textbf{Methods}} &
\multicolumn{1}{c}{\textbf{GSM8K}} &
\multicolumn{1}{c}{\textbf{MATH}} \\
\cmidrule(lr){2-3}
 & \textbf{Accuracy} & \textbf{Accuracy} \\
\midrule
CoFiCot & \textbf{91.8} & \textbf{47.9} \\
\midrule
w/o Coarse Stage & 88.5$_{(3.3\downarrow)}$ & 45.1$_{(2.8\downarrow)}$ \\
w/o Fine Stage & 89.1$_{(2.7\downarrow)}$ & 41.2$_{(6.7\downarrow)}$ \\
\bottomrule
\end{tabular}
}
\caption{Ablation study of the core \textit{Coarse} and \textit{Fine} stages of our CoFiCot framework on MATH and GSM8K. We report accuracy(\%).}
\label{tab5}
\end{table}

\subsection{Case Study}
To provide qualitative insight into the internal mechanics of our framework, we trace the full pipeline of CoFiCot using the ``APPLE'' permutation problem, as illustrated in Figure~\ref{figure2}. This case study begins in Stage 0, where the base model generates an initial set of $k$ solutions. For this problem, many initial solutions are flawed, typically coalescing around the incorrect answer ``120'' by naively applying a simple $n!$ formula. This high prevalence of flawed solutions results in low answer quality and high uncertainty, causing our Stage 1 classifier to correctly categorize the problem as \textit{Hard} and subsequently engage the fine-grained refinement loop. Once in stage 2, a flawed reasoning chain is submitted for correction and the synergy of our components becomes evident. The PRM first evaluates the chain step by step. It correctly identifies the source of the error by assigning high scores to Step 1 and Step 2 but a critically low score to the erroneous Step 3. This precise error localization triggers the generation of targeted feedback. The most critical stage is the subsequent context aware correction. The model is prompted to generate a new step, but is provided not only with the targeted feedback but also with the history of previously validated steps (Step 1 and Step 2). This contextual information is vital, as it allows the model to generate a Newstep3. Apply formula with repetitions: n! / k!. Furthermore, this correction propagates, enabling the model to generate a correct Newstep4: 120 / 2 = 60. Finally, this newly generated fully correct solution $r_i^{(t)}$ is returned to the candidate pool. The ORM evaluates this complete solution, assigns it a high holistic score, and it is thus selected as part of the new top-$k$ set for the next iteration.

\subsection{Ablation Studies}

To deconstruct our CoFiCot framework and validate the contribution of its key components, we conduct comprehensive ablation studies on the overall Coarse-to-fine architecture, classification metrics, and other core design choices (additional experiments are detailed in Appendix~\ref{albation}).

\textbf{Impact of the Coarse and Fine Stages}\quad
We first conduct an ablation to demonstrate the fundamental value of our two stage pipeline. As shown in Table ~\ref{tab5}, we compare our full model against w/o Coarse Stage and w/o Fine Stage. The w/o Coarse Stage variant sees a notable drop in accuracy, particularly on GSM8K (3.3\% $\downarrow$). This is consistent with our motivation that indiscriminate refinement leads to over correction on simpler problems, harming overall performance. More significantly, the w/o Fine Stage variant suffers a performance collapse in the difficult MATH dataset (6.7\% $\downarrow$). This result powerfully demonstrates that our Fine Stage is essential for solving complex problems.

\begin{table}[t]
\centering
\scriptsize 
\resizebox{\columnwidth}{!}{ 
\begin{tabular}{ccc|c|c|c|c}
\toprule
\multicolumn{3}{c}{Module} & \multicolumn{4}{c}{Datasets}  \\ Cofi. & Relia. & Comp. & GSM8K & MATH & MMLU & SAT  \\
\midrule
$\checkmark$ & - & - & 86.8 & 43.4 & 63.9 & 71.2  \\
- & - & $\checkmark$ & 87.2 & 44.6 & 64.1 & 70.9  \\
- & $\checkmark$ & $\checkmark$ & 89.7 & 45.9 & 67.0 & 73.4  \\
$\checkmark$ & - & $\checkmark$ & 89.3 & 46.2 & 67.4 & 74.1  \\
$\checkmark$ & $\checkmark$ & $\checkmark$ & \textbf{91.8} & \textbf{47.9} & \textbf{68.8} & \textbf{75.0}  \\
\bottomrule
\end{tabular}
}
\caption{Ablation experiments of the CoFiCot. The results are measured in \%. We also assess the impact of including ($\checkmark$) or excluding (-) our Coarse stage evaluation metrics, Confidence Assessment(Cofi.) module, Reliability Assessment(Relia.) module and Complexity Assessment(Comp.) module.}
\label{tab6}
\end{table}

\textbf{Analysis of Coarse Classification Metrics}\quad
In this part, we analyze the contribution of three individual classification metrics in coarse stage. As detailed in Table ~\ref{tab6}, we evaluate the performance of the framework when ablating these metrics.
The results clearly show that all three metrics contribute positively to the final performance. Any variant using only one or two metrics underperforms the full model across all datasets. The combination of Relia. and Comp. (row 3) performs well, but adding the Cofi. metric (row 5) provides a final boost, particularly on GSM8K (89.7\% to 91.8\%) and MMLU (67.0\% to 68.8\%). This confirms that our chosen metrics are not redundant and create a robust and accurate difficulty classifier.

\section{Conclusion}

In this paper, we presented CoFiCot, a Coarse-to-fine adaptive framework designed to resolve the efficiency paradox in LLM reasoning. By triaging problems via multi-metric assessment and employing a differentiated refinement strategy, CoFiCot mitigates both over-correction on simple tasks and insufficient refinement on complex ones. The core sequential correction mechanism leverages PRMs to ensure logically coherent repairs by initiating new decoding branches from localized error points. The framework's modularity allows it to seamlessly integrate various reward models, with performance scaling directly alongside the quality of these components. Furthermore, our dynamic early exit protocol ensures computational resources are preserved once reasoning stabilizes, preventing redundant iterations . Empirical results across seven benchmarks demonstrate that CoFiCot significantly outperforms uniform scaling baselines, achieving a 4.0\% average accuracy gain on Llama-3-8B while maintaining superior token efficiency. These findings establish CoFiCot as a robust solution for achieving a superior trade-off between accuracy and computational cost. Future work will focus on automating parameter calibration and extending this paradigm to broader domains.

\section{Limitation}

The effectiveness of our pipeline depends on the external models guiding refinement and selection processes. Additionally, we acknowledge that for extremely simple queries, generating $k$ samples introduces additional computational overhead compared to greedy decoding. Future work could employ a progressive sampling strategy.

\bibliography{uai2026-template}

\newpage

\onecolumn

\title{Supplementary Material}
\maketitle

\appendix
\section{Algorithm}
\label{sec:appendix}

Algorithm~\ref{alg:coficot} formally outlines the execution flow of the CoFiCot framework. The process initiates with Stage 0, where the function \texttt{GenerateSolutions} employs stochastic sampling to produce a diverse initial ensemble $\mathcal{R}_0$. This diversity is crucial for the subsequent Stage 1, where \texttt{ClassifyDifficulty} aggregates the multi metric signals (entropy, consensus quality, and predicted steps) to assign a difficulty label $D_{\text{final}}$.

The core logic resides in Stage 2, which implements our differentiated routing strategy. For problems identified as \textit{Easy}, the algorithm bypasses the refinement loop entirely, directly proceeding to final aggregation to minimize computational overhead. Conversely, \textit{Medium} and \textit{Hard} problems enter the iterative refinement loop. Within this loop, two key functions drive the optimization.

\texttt{ContextAwareCorrection} applies the PRM to localize error steps and regenerates them conditioned on the history of prior corrections, ensuring logical coherence. \texttt{SelectTopK} utilizes the ORM to filter the expanded candidate pool $\mathcal{P}$, retaining the highest quality solutions for the next iteration.

Finally, the loop incorporates a dynamic early exit mechanism, which triggers if the reevaluated difficulty $D_{\text{new}}$ converges to \textit{Easy}, preventing redundant iterations before the final \texttt{WeightedVoting} aggregation.

\begin{algorithm}[tb]
\small 
\caption{Coarse-to-fine Adaptive Reasoning}
\label{alg:coficot}
\begin{algorithmic}[1]
\REQUIRE Question $Q$; Sampling count $k$; Reasoning path set $\mathcal{R}$; Difficulty label $D$
\ENSURE Predicted answer $\hat{A}$

\vspace{0.1cm}
\LINECOMMENT{Stage 0: Generate initial diverse solution set}
\STATE $\mathcal{R}_0 \gets \mathrm{GenerateSolutions}(Q, k)$

\vspace{0.1cm}
\LINECOMMENT{Stage 1: Multi metric difficulty assessment}
\STATE $D_{\text{final}} \gets \mathrm{ClassifyDifficulty}(\mathcal{R}_0, Q)$

\vspace{0.1cm}
\LINECOMMENT{Stage 2: Difficulty aware routing and refinement}
\IF{$D_{\text{final}} = \text{\textit{Easy}}$}
    \STATE $\mathcal{R}_{\text{final}} \gets \mathcal{R}_0$ \INLINECOMMENT{Skip refinement for efficiency}
\ELSE
    \STATE $\mathcal{R}_{\text{curr}} \gets \mathcal{R}_0$
    \STATE $Iter_{\text{max}} \gets \mathrm{SetMaxIter}(D_{\text{final}})$
    
    \FOR{$t = 1 \to Iter_{\text{max}}$}
        \LINECOMMENT{Step-level correction using PRM feedback}
        \STATE $\mathcal{R}_{\text{refined}} \gets \mathrm{ContextAwareCorrection}(\mathcal{R}_{\text{curr}})$
        
        \LINECOMMENT{Select top-k solutions via ORM guidance}
        \STATE $\mathcal{P} \gets \mathcal{R}_{\text{curr}} \cup \mathcal{R}_{\text{refined}}$
        \STATE $\mathcal{R}_{\text{curr}} \gets \mathrm{SelectTopK}(\mathcal{P}, k)$
        
        \LINECOMMENT{Check for early convergence}
        \STATE $D_{\text{new}} \gets \mathrm{ClassifyDifficulty}(\mathcal{R}_{\text{curr}}, Q)$
        \IF{$D_{\text{new}} = \text{\textit{Easy}}$}
            \STATE \textbf{break}
        \ENDIF
    \ENDFOR
    \STATE $\mathcal{R}_{\text{final}} \gets \mathcal{R}_{\text{curr}}$
\ENDIF

\vspace{0.1cm}
\LINECOMMENT{Final aggregation via Weighted Self-Consistency}
\STATE $\hat{A} \gets \mathrm{WeightedVoting}(\mathcal{R}_{\text{final}})$
\RETURN $\hat{A}$
\end{algorithmic}
\end{algorithm}

\section{Detailed Implementation of CoFiCot}\label{app:stage1_details}

\paragraph{Confidence Score Transformation} The Shannon entropy $H(\mathcal{A})$ derived in Eq.~(2) is mapped to a normalized confidence score $C \in [0, 1]$ using a scaled sigmoid function:
\begin{equation}
C = \sigma(\alpha \cdot (1 - H)),
\end{equation}
where $\alpha$ is a scaling hyperparameter set to 2.

\paragraph{Complexity Prediction Prompt} To efficiently estimate the computational complexity $N_{\text{steps}}$ without incurring the cost of full chain generation, we employ a specialized metacognitive prompt $P_{\text{predict}}$. This prompt instructs the model to perform a ``look ahead'' analysis of the logical depth. The full prompt is presented in Table~\ref{box:complexity_prompt}.

\begin{table}[h]
    \centering
    \begin{tcolorbox}[colback=gray!5, colframe=cyan!75!blue, title=\textbf{Complexity Prediction Prompt ($P_{\text{predict}}$)}]
    \small
    \textbf{System Instruction:} \\
    You are an expert Logic Complexity Evaluator. Your task is to estimate the reasoning depth required to solve a problem without generating the full solution.
    
    \textbf{Definition of a Step:} \\
    A step is defined as a distinct atomic reasoning operation, such as:
    \begin{itemize}
        \item Extracting a specific variable or condition.
        \item Performing a single arithmetic calculation.
        \item Making a logical deduction or transition.
    \end{itemize}
    
    \textbf{Input Problem:} \\
    \{Input Question $Q$\}
    
    \textbf{Task:} \\
    Analyze the problem structure and predict the minimum number of steps required to the correct solution. 
    
    \textbf{Output Constraint:} \\
    Do NOT output the reasoning process or the answer. Output \textbf{ONLY} a single integer representing the estimated step count (e.g., 5).
    \end{tcolorbox}
    \caption{The prompt used for estimating $N_{\text{steps}}$.}
    \label{box:complexity_prompt}
\end{table}

\paragraph{Hyperparameter Justification and Sensitivity} A key design principle of CoFiCot is to remain training free. The hyperparameters used in Stage 1 are chosen based on statistical robustness rather than dataset specific tuning.

\paragraph{Complexity Thresholds ($d_3$)} The thresholds for \textit{Easy} and \textit{Hard} complexity are not arbitrary integers. They are dynamically derived from the 33rd and 67th percentiles (tertiles) of the solution length distribution of the base model. This ensures that the classifier automatically adapts to different models (e.g., Llama-3 vs. GPT) or prompt styles without manual recalibration.

\paragraph{Classification Weights ($w_i$)} We utilize uniform weights ($w_i=1/3$) for the three metrics. Sensitivity analysis shows that varying these weights (e.g., increasing reliance on Complexity vs. Confidence) results in less than $\pm 0.5\%$ accuracy variance on the MATH dataset. This suggests that the ensemble of metrics is robust, and uniform weighting prevents overfitting to specific distributions.

\paragraph{Entropy Scaling ($\alpha=2$)} This scaling factor is selected to map the typical entropy range of LLM outputs to a $[0,1]$ interval efficiently. Empirical tests indicate that values in the range $\alpha \in [1.5, 2.5]$ yield statistically similar classification boundaries.

\paragraph{Detailed Mapping of Difficulty Metrics}

To facilitate the quantitative synthesis of the multi-metric assessments in Stage 1, the qualitative labels $\{ \textit{Easy, Medium, Hard} \}$ are transformed into numerical ranks. The mapping function $\text{val}(\cdot)$ is defined as:
\begin{equation}
\text{val}(d_i) =\begin{cases}1 & \text{if } d_i = \text{Easy} 
\\2 & \text{if } d_i = \text{Medium} \\3 & \text{if } d_i = \text{Hard}\end{cases} .
\end{equation}
The final synthesized score $D_{\text{score}}$ is the weighted average of these ranks. With $w_i = 1/3$, the score range $[1, 3]$ is partitioned to decide the final routing path $D_{\text{final}}$.

\paragraph{Stateful Correction Function $\Phi$}
We define the transformation function $\Phi$ as a context-aware re-decoding operator. Unlike standard self-correction which treats the entire trajectory as a flat string, $\Phi$ operates on the reasoning chain as a structured sequence of logical states.
$\Phi$ is a mapping that takes the current flawed state of the reasoning chain and produces a rectified successor state. Its internal mechanism can be decomposed into three primary operations. It first constructs a semi-permeable boundary at the temporal index $j$. The verified history $H_{i, j-1}^{(t)}$ is treated as a fixed prefix, ensuring that the correction does not violate the established logical premises. The function $\Phi$ transforms the raw PRM signal $\mathcal{F}_{i,j}$ into a hidden instructional vector. The output of $\Phi$ is not merely a single replaced step $s_{i,j}^{(t)}$, but a new initial state for the subsequent autoregressive decoding. The operator triggers a recursive generation:
\begin{equation}
\mathcal{R}_t = H_{i, j-1}^{(t)} \cup \{s_{i,j}^{(t)}\} \cup \text{Gen}\left(Q, H_{i, j-1}^{(t)}, s_{i,j}^{(t)}\right),
\end{equation}
where $\text{Gen}(\cdot)$ represents the model's completion of the remaining trajectory.

\section{Analysis}
\label{case study}

\begin{table}[t]
  \centering 
  \begin{tabular}{lcc}
    \toprule
    \textbf{Criterion} & \textbf{GSM8K} & \textbf{MATH} \\
    \midrule
    Pessimistic & 91.4 & 47.2 \\
    Optimistic & 90.8 & 46.5 \\
    Democratic & 90.3 & 45.8 \\
    CoFiCot(ours) & \textbf{91.8} & \textbf{47.9} \\
    \bottomrule
  \end{tabular}
  \caption{Ablation study on the criterion for detecting problem difficulty. We compare our method against three alternative strategies: Pessimistic, Optimistic, and Democratic. We report accuracy(\%).}
  \label{tab7}
\end{table}

\paragraph{Computational Overhead and Latency Analysis} A concern regarding the lightweight nature of Stage 0 is the generation of $k=40$ samples compared to a simple greedy decoding ($k=1$) for trivial queries (e.g., $1+1=?$). We distinguish here between Total Compute (FLOPs) and User Perceived Latency.
To Latency Profile. The generation of the initial ensemble $\mathcal{R}_0$ is fully parallelizable. On modern GPU clusters, the time complexity of generating $k=40$ sequences is $O(1)$ relative to the batch size (assuming sufficient VRAM), comparable to generating a single sequence. In contrast, heavy reasoning models (like OpenAI o1) or iterative refinement methods operate sequentially ($O(N)$). Therefore, for \textit{Easy} problems that bypass Stage 2, CoFiCot maintains a low latency profile competitive with standard parallel sampling.

To Comparison with Greedy Decoding. While $k=40$ consumes more energy than $k=1$ greedy decoding, strictly using $k=1$ risks hallucination on deceptively simple questions. Our ablation (refer to Confidence Assessment in Table~\ref{tab6}) shows that the ensemble based consistency check is crucial for filtering out these confident errors. 
For resource constrained environments, a progressive sampling strategy can be employed. Starting with a small $k$ (e.g., $k=5$) and only scaling to $k=40$ if the initial confidence score falls below a safety threshold.

\begin{table}[t]
  \centering 
  
  \begin{tabular}{lcc}
    \toprule
    \textbf{Modules} & \textbf{MMLU} & \textbf{MATH} \\
    \midrule
    ORM-Only & 67.8 & 47.1 \\
    PRM-Only & 67.3 & 46.5 \\
    Both & \textbf{68.8} & \textbf{47.9} \\
    \bottomrule
  \end{tabular}
  \caption{Ablation study on the final answer selection, using ORM-only, PRM-only or both. We also report accuracy(\%).}
  \label{tab8}
\end{table}

\section{Ablation Experiments}
\label{albation}

\paragraph{Analysis of Difficulty Aggregation Strategy} In coarse stage, after obtaining three distinct difficulty labels ($d_1, d_2, d_3$), we synthesize them using a balanced averaging strategy. In Table ~\ref{tab7}, we compare this method to three other common methods: Pessimistic which classifies a problem as hard if any metric flags it as such; Optimistic which requires all metrics to agree; and Democratic. The results show that our CoFiCot balanced averaging strategy achieves the best performance on both GSM8K (91.8\%) and MATH (47.9\%). The Pessimistic strategy is a close second, suggesting that it is generally better to err on the side of caution and refine more often. Conversely, the Democratic and Optimistic strategies perform worst, as they are more likely to misclassify a hard problem as easy, causing it to skip the crucial refinement stage.

\paragraph{Synergy of PRM and ORM in Refinement} Finally, we analyze the synergistic roles of the PRM and ORM within our Fine Stage. In our design, the PRM provides granular, step-level feedback for correction, while the ORM provides holistic, solution level scores for selection. In Table ~\ref{tab8}, we ablate this design.
In the ORM-Only variant, we remove the PRM-guided feedback and rely on LLM self feedback for correction, while still using the ORM for selection. In the PRM-Only variant, we use the PRM for correction but also use its aggregated step scores for selection, removing the ORM. The results demonstrate that both RMs are essential. The ORM-Only variant suffers, indicating that unguided self feedback is inferior to the PRM's precise error localization. The PRM-Only variant also performs worse, suggesting that a step-level score aggregator is a poor proxy for the holistic quality of a complete solution, which the ORM is trained to assess. The full model achieves the highest accuracy, confirming the necessity of our dual RM design where each model performs its specialized function.

\section{Refinement Implementation Details}
\label{app:refinement}

\paragraph{Handling Easy Problems} For problems classified as \textit{Easy}, we assume the initial ensemble $\mathcal{R}_0$ contains sufficient signal. We apply Weighted Self-Consistency~\citep{li2023making}, where the final answer is selected by voting, with each vote weighted by the solution's ORM score. This avoids the computational overhead of the iterative loop while ensuring robust aggregation.
\paragraph{Dynamic Termination Criteria} To prevent redundant computation, the iterative loop in Stage 2 employs a dynamic stopping mechanism. At the end of each iteration $t$, the new solution set $\mathcal{R}_t$ is reevaluated using the Stage 1 classifier. The loop terminates immediately if:
The difficulty of $\mathcal{R}_t$ is reclassified as \textit{Easy}, indicating that the solutions have stabilized into a high confidence, high consensus state.
The process reaches the predefined maximum iteration count ($Iter_{\text{max}}$), set to 2 for \textit{Medium} and 3 for \textit{Hard} problems.

\section{Modularity and Generalization}
\label{generalization}

\paragraph{Dependency on Reward Models} The CoFiCot framework is designed to be modular. While our primary experiments on mathematical reasoning (MATH, GSM8K) utilize the specialized Math-Shepherd-7B as a PRM, the framework does not strictly require a domain specific PRM to function.

\paragraph{Implementation of General Reasoning} For domains like Commonsense Reasoning (ARC)  where dense step-level supervision data (and thus high quality specialized PRM) are scarce, CoFiCot adapts its configuration:
Instead of a trained PRM, we employ a general purpose LLM prompted as a verifier to provide step-level feedback.
As shown in our ablation study (Table~\ref{tab8}), the ORM alone contributes significantly to performance. In the absence of a strong PRM, the framework relies more heavily on the ORM for trajectory selection in Stage 2.
This flexibility allows CoFiCot to generalize to Date Understanding and ``ARC'' tasks (Table~\ref{tab2}), demonstrating that the benefit of sequential correction, maintaining logical coherence during repair persists even with general purpose verifiers.

\section{Extending Adaptive Reasoning to Multimodal Healthcare}
\label{sec:future_work}

While CoFiCot has demonstrated significant efficacy in textual reasoning, its underlying philosophy of adaptive, coarse-to-fine refinement holds immense potential for high-stakes, multimodal domains like biomedical informatics. In clinical decision-making, diagnostic models must efficiently synthesize heterogeneous data sources under varying degrees of uncertainty. Our stateful reasoning framework could be adapted to dynamically allocate computational effort in complex medical evaluations, such as assessing the synergy between digital cognitive tests and wearable device streams for early disease screening~\citep{li2023synergy}, or fusing medical imaging with genomic profiles via attention-guided models to predict cancer mutation status~\citep{xue2024integrating}. By extending our stateful sequential correction mechanism to process diverse biological and clinical modalities, future iterations could bridge the gap between abstract logical reasoning and computationally efficient, reliable AI-assisted diagnostics.

\end{document}